\documentclass[pdflatex,sn-nature]{sn-jnl} 

\usepackage{graphicx}
\usepackage{amsmath,amssymb,amsfonts}
\usepackage{amsthm}
\usepackage{xcolor}
\usepackage{textcomp}
\usepackage{booktabs}
\usepackage{url}

\theoremstyle{thmstyleone}

\theoremstyle{thmstyletwo}

\theoremstyle{thmstylethree}

\begin{document}

\title[Energy-conditioned flow matching for adsorbate placement]{AdsorbFlow: energy-conditioned flow matching enables fast and realistic adsorbate placement }

\author[1]{\fnm{Jiangjie} \sur{Qiu}}
\author[1]{\fnm{Wentao} \sur{Li}}
\author[1]{\fnm{Honghao} \sur{Chen}}
\author[1]{\fnm{Leyi} \sur{Zhao}}
\author*[1]{\fnm{Xiaonan} \sur{Wang}}\email{wangxiaonan@tsinghua.edu.cn}

\affil[1]{\orgname{Beijing Key Laboratory of Artificial Intelligence for Advanced Chemical Engineering Materials, Department of Chemical Engineering, Tsinghua University},
\orgaddress{\city{Beijing}, \country{China}}}

\abstract{
Identifying low-energy adsorption geometries on catalytic surfaces is a practical bottleneck for computational heterogeneous catalysis: the difficulty lies not only in the cost of density functional theory (DFT) but in proposing initial placements that relax into the correct energy basins. Conditional denoising diffusion has improved success rates, yet requires $\sim$100 iterative steps per sample.

Here we introduce \textit{AdsorbFlow}, a deterministic generative model that learns an energy-conditioned vector field on the rigid-body configuration space of adsorbate translation and rotation via conditional flow matching. Energy information enters through classifier-free guidance conditioning---not energy-gradient guidance---and sampling reduces to integrating an ODE in as few as \textbf{5} steps.

On OC20-Dense with full DFT single-point verification, AdsorbFlow with an EquiformerV2 backbone achieves \textbf{61.4\%} SR@10 and \textbf{34.1\%} SR@1---surpassing AdsorbDiff (31.8\% SR@1, 41.0\% SR@10) at every evaluation level and AdsorbML (47.7\% SR@10)---while using $20{\times}$ fewer generative steps and achieving the lowest anomaly rate among generative methods (6.8\%). On 50 out-of-distribution systems, AdsorbFlow retains \textbf{58.0\%} SR@10 with a MLFF-to-DFT gap of only 4~percentage points. These results establish that deterministic transport is both faster and more accurate than stochastic denoising for adsorbate placement.
}

\maketitle

\section*{Introduction}

The structure and energetics of adsorbed intermediates at solid surfaces govern the activity and selectivity of heterogeneous catalysts, underpinning large fractions of chemical manufacturing and emerging energy-conversion technologies. In silico screening using density functional theory (DFT) has become indispensable for exploring this design space, yet the practical throughput of such workflows remains constrained by a deceptively mundane step: \emph{placing} an adsorbate on a slab in a way that relaxes to the correct low-energy adsorption basin.

The Open Catalyst 2020 (OC20) and its dense-placement extension (OC20-Dense) crystallized the difficulty of this global-minimum search problem by providing standardized data and end-to-end evaluation pipelines that combine machine-learned force-field (MLFF) relaxation with DFT verification \cite{Chanussot2021OC20,Lan2023AdsorbML}. AdsorbML showed that brute-force enumeration of many initial sites can recover near-optimal minima, but at the cost of many relaxations per system \cite{Lan2023AdsorbML}. More recently, AdsorbDiff proposed a conditional denoising diffusion model over 2D translations and 3D rigid rotations, conditioned on \emph{relative adsorption energies} of dense local minima to improve single-site success rates \cite{Kolluru2024AdsorbDiff}. However, diffusion-based placement typically requires nearly 100 reverse steps per sample, limiting throughput in high-volume screening campaigns.

A key observation motivates our approach: the physical relaxation from an initial placement toward a stable adsorption geometry is fundamentally deterministic---practitioners ultimately run deterministic optimizers (e.g., L-BFGS) on a learned or DFT energy landscape. The mismatch between deterministic downstream evaluation and stochastic generative sampling is both conceptually unsatisfying and practically wasteful: it introduces unnecessary variance and forces the generator to invert a noise schedule rather than learn a direct transport map.

In this work we introduce \textit{AdsorbFlow}, which replaces stochastic reverse diffusion with \emph{energy-conditioned deterministic transport}. We learn a time-dependent vector field on the rigid-body configuration manifold of translation and rotation, trained by conditional flow matching \cite{Lipman2023FlowMatching} with rectified-flow paths that explicitly straighten transport trajectories \cite{Liu2022RectifiedFlow}. Energy enters through \emph{classifier-free guidance conditioning} \cite{Ho2022CFG}: relative energies are provided during training, randomly dropped to learn a paired unconditional model, and combined at inference to focus sampling toward low-energy basins. Crucially, AdsorbFlow requires neither an explicit differentiable energy model nor $\nabla_x E$.

Our contributions are twofold:
\begin{enumerate}
    \item \textbf{A generative paradigm shift.} We show that deterministic flow matching achieves DFT success rates of 34.1\% (SR@1) and 61.4\% (SR@10) with an EquiformerV2 backbone on OC20-Dense, using only 5 generative steps---a $20{\times}$ reduction from the ${\sim}$100 steps of AdsorbDiff---surpassing AdsorbDiff (31.8\% SR@1, 41.0\% SR@10) and AdsorbML (9.1\% SR@1, 47.7\% SR@10) at every evaluation level while achieving the lowest anomaly rate among generative methods. On a 50-system out-of-distribution split, AdsorbFlow retains 58.0\% SR@10, confirming robust generalization.
    \item \textbf{Backbone-agnostic design with principled rotation heads.} We show that AdsorbFlow naturally accommodates both lightweight ($\ell{\le}1$, PaiNN) and higher-order ($\ell{\le}4$, EquiformerV2) equivariant backbones through principled rotation-head designs---torque aggregation for PaiNN vs.\ direct pseudovector regression for EquiformerV2---and that backbone expressivity and CFG guidance strength interact non-trivially.
\end{enumerate}

\section*{Results}

\subsection*{Adsorbate placement as transport on a symmetry-constrained configuration space}

We consider an adsorbate--slab system with slab atoms $\mathcal{S}$, adsorbate atoms $\mathcal{A}$, and a set of locally minimized adsorption geometries with energies provided by OC20-Dense \cite{Lan2023AdsorbML}. Following the community-standard protocol established by AdsorbDiff \cite{Kolluru2024AdsorbDiff}, we model the adsorbate as a rigid body undergoing in-plane translation and SO(3) rotation:
\begin{equation}
x = (\mathbf{t}, R), \quad \mathbf{t}=(t_x,t_y)\in\mathbb{T}^2,\; R\in\mathrm{SO}(3),
\end{equation}
where $\mathbf{t}$ is the in-plane translation of the adsorbate center of mass (COM) on a 2-torus under periodic boundary conditions, and $R$ is a rotation applied about the COM. The vertical coordinate $z$ is not modeled by the generative process; instead, it is determined by subsequent MLFF relaxation, which naturally adjusts the adsorbate height to accommodate the local surface geometry.

While this rigid-body assumption is highly effective for small-to-medium adsorbates, it is important to note its boundary conditions: for industrially relevant large and complex molecules with internal torsional degrees of freedom, treating the adsorbate as a rigid body may lead to insufficient conformational exploration.

\subsection*{AdsorbFlow: energy-conditioned flow matching with classifier-free guidance}

AdsorbFlow learns a time-dependent vector field
\begin{equation}
v_\theta(x,t \mid c):\; (x,t,c)\mapsto (\dot{\mathbf{t}},\boldsymbol{\omega}),
\end{equation}
where $\boldsymbol{\omega}\in\mathbb{R}^3$ is an angular velocity in the Lie algebra $\mathfrak{so}(3)$ and $c$ denotes conditioning information including (i) the slab/adsorbate atomic graph and (ii) a scalar \emph{relative energy condition} $E_{\mathrm{rel}}$.

\begin{figure}[t]
    \centering
    \includegraphics[width=\linewidth]{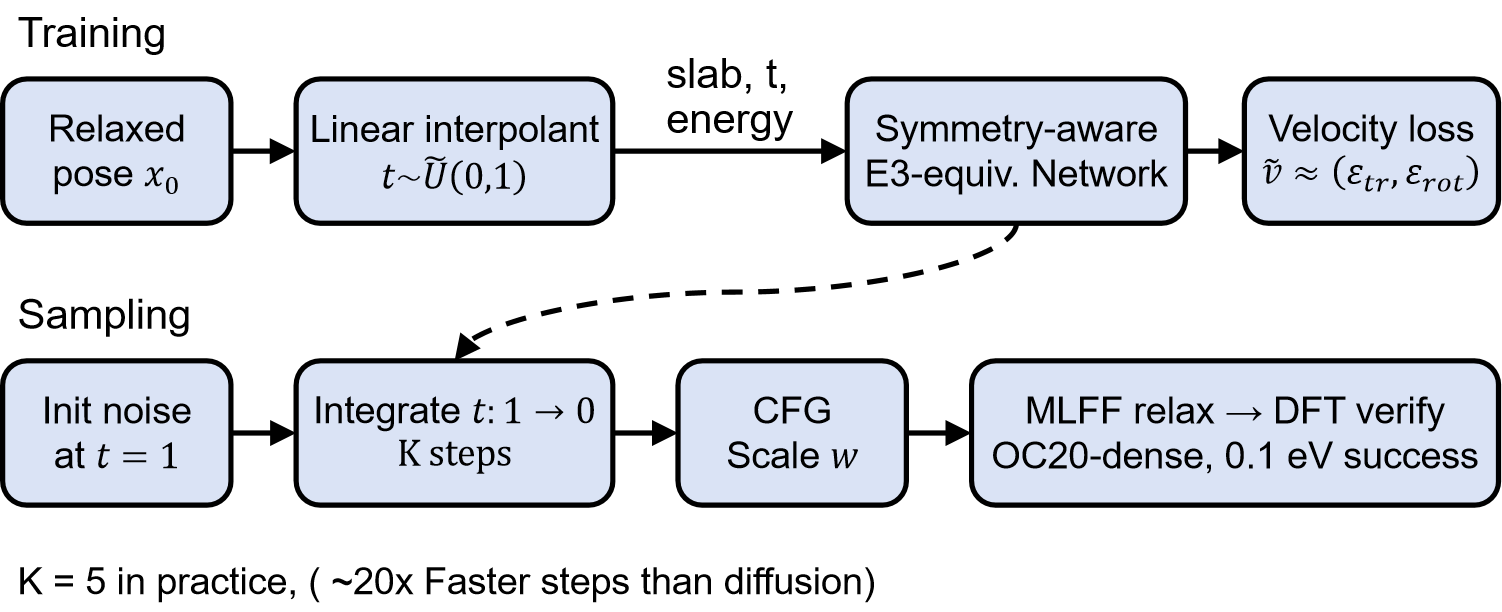}
    \caption{\textbf{AdsorbFlow overview.} \textbf{Training:} A linear interpolant connects the relaxed pose $x_0$ (at $t{=}0$) to noise $x_1$ (at $t{=}1$), and a symmetry-aware E(3)-equivariant network predicts the velocity field. \textbf{Sampling:} Starting from initial noise $x_1$ at $t{=}1$, we integrate the ODE backward to $t{=}0$ using a Heun solver for $K$ steps (default $K{=}5$), applying classifier-free guidance (CFG). The generated placements are then relaxed via MLFF and verified with DFT.}
    \label{fig:overview}
\end{figure}

Sampling is deterministic: we draw $x_1$ from a simple prior (uniform in-plane position and random rotation) and integrate the ODE backward from $t{=}1$ to $t{=}0$:
\begin{equation}
\frac{d\mathbf{t}}{dt}=\dot{\mathbf{t}},\qquad
\frac{dR}{dt}=\left[\boldsymbol{\omega}\right]_\times R,
\end{equation}
for a fixed number of steps $K$ (default $K=5$). Here $[\cdot]_\times$ denotes the skew-symmetric matrix associated with a 3D vector. An overview of the full pipeline is shown in Fig.~\ref{fig:overview}.

\textbf{Conditional flow matching on translation and rotation.}
We train $v_\theta$ with a flow-matching objective constructed from pairs $(x_0,x_1)$ where $x_0$ is a relaxed local minimum (at $t{=}0$) and $x_1$ is a randomly perturbed rigid-body placement drawn from the prior (at $t{=}1$). For translation, we use a linear interpolant with PBC-aware displacement. For rotation, we use geodesic interpolation on $\mathrm{SO}(3)$:
\begin{equation}
R_t = \exp\!\left(t\,\log(R_1 R_0^\top)\right)R_0,
\end{equation}
with the linear schedule $\beta(t){=}t$ and matrix log/exp on $\mathrm{SO}(3)$, ensuring endpoints $R_{t=0}{=}R_0$ and $R_{t=1}{=}R_1$. This left-multiply geodesic form is consistent with the space-frame ODE convention $\tfrac{dR}{dt}{=}[\boldsymbol{\omega}]_\times R$. The corresponding target angular velocity is $\boldsymbol{\omega}^\ast{=}\log(R_1 R_0^\top)^\vee$, where $(\cdot)^\vee\colon\mathfrak{so}(3)\to\mathbb{R}^3$ denotes the inverse of the hat map (i.e., extracting the 3-vector from a skew-symmetric matrix). Because $\beta(t){=}t$, this velocity is constant along each interpolant path, consistent with the rectified-flow objective. The training loss regresses the vector field to these targets:
\begin{equation}
\mathcal{L}=\mathbb{E}\Big[\|\dot{\mathbf{t}}_\theta-\dot{\mathbf{t}}^\ast\|^2 + \lambda_{\mathrm{rot}}\|\boldsymbol{\omega}_\theta-\boldsymbol{\omega}^\ast\|^2\Big].
\end{equation}

\textbf{Energy conditioning via CFG (not energy-gradient guidance).}
Each OC20-Dense system provides $\sim$100 local minima with energies. We compute a per-system relative energy
\begin{equation}
E_{\mathrm{rel}} = E - E_{\min}.
\end{equation}
An energy embedding network maps $E_{\mathrm{rel}}$ to FiLM \cite{Perez2018FiLM} parameters that modulate scalar node features in an equivariant GNN backbone (PaiNN by default). During training, we randomly drop the energy condition with probability $p_{\mathrm{cfg}}$, forcing the model to learn both conditional and unconditional vector fields. At inference, we can optionally apply classifier-free guidance by combining the two predictions:
\begin{equation}
v_{\mathrm{cfg}} = (1+w)\,v_\theta(\cdot\mid E_{\mathrm{rel}}) - w\,v_\theta(\cdot\mid \varnothing),
\end{equation}
with guidance scale $w\ge 0$. Importantly, this is \emph{conditioning-based} guidance; AdsorbFlow does not require an explicit energy model nor $\nabla_x E$.

\subsection*{AdsorbFlow achieves strong placement success with 5 generative steps}

We evaluate AdsorbFlow on the OC20-Dense benchmark (44 in-distribution systems) following the end-to-end evaluation pipeline of AdsorbDiff \cite{Lan2023AdsorbML,Kolluru2024AdsorbDiff}. For each system, we generate 10 candidate placements with independent random seeds. Each placement is relaxed using a pretrained GemNet-OC MLFF \cite{Gasteiger2022GemNetOC} and screened for geometric anomalies (desorption, dissociation, slab reconstruction, intercalation). Following AdsorbDiff's convention, for SR@$k$, the non-anomalous candidate with the lowest MLFF energy among seeds $0,\ldots,k{-}1$ undergoes one VASP DFT single-point verification, and a system is successful if the verified adsorption energy lies within 0.1~eV of the best-known DFT reference minimum \cite{Kolluru2024AdsorbDiff}.

Table~\ref{tab:main} summarizes the main results. We select hyperparameters via MLFF-level grid search over guidance scale $w\in\{0,1,3,5,7,10\}$ and integration steps $K\in\{5,10,30\}$, then verify the best configuration with full DFT (664 VASP single-point calculations across both backbones). With an EquiformerV2 backbone at $w{=}7$ and only \textbf{5} generative steps, AdsorbFlow achieves a DFT SR@1 of \textbf{34.1\%} and SR@10 of \textbf{61.4\%}. With a lightweight PaiNN backbone at $w{=}5$ and $K{=}5$, AdsorbFlow achieves 27.3\% and 47.7\%. Both require only 5 forward passes of the equivariant backbone per candidate---a $20{\times}$ reduction from the ${\sim}$100 denoising steps used by diffusion baselines.

\begin{table}[t]
\centering
\caption{\textbf{OC20-Dense placement success (DFT-verified, 44 ID systems).} SR@$k$: for each system, the non-anomalous candidate with the lowest MLFF energy among the first $k$ seeds is selected and verified by a single DFT calculation; a system is successful if the verified energy lies within 0.1~eV of the DFT reference minimum. Although DFT single-point energies were computed for all non-anomalous candidates to enable post-hoc analysis, SR@$k$ selection uses only MLFF energies; DFT is used strictly for verification. Anom.@10: fraction of systems with all 10 candidates anomalous. ``---'' indicates values not reported. AdsorbDiff baseline from \cite{Kolluru2024AdsorbDiff}; AdsorbML baseline from \cite{Lan2023AdsorbML}.}
\label{tab:main}
\begin{tabular}{l l c c c c c c}
\toprule
Method & Backbone & Steps & \multicolumn{4}{c}{DFT SR (\%)} & Anom.@10 \\
\cmidrule(lr){4-7}
 & & & @1 & @2 & @5 & @10 & (\%) \\
\midrule
AdsorbML & --- & --- & 9.1 & 20.5 & 34.1 & 47.7 & 6.8 \\
AdsorbDiff & EqV2 & ${\sim}$100 & 31.8 & 34.1 & 36.3 & 41.0 & 13.6 \\
\midrule
\textbf{AdsorbFlow} & EqV2 & \textbf{5} & \textbf{34.1} & \textbf{45.5} & \textbf{54.5} & \textbf{61.4} & \textbf{6.8} \\
\textbf{AdsorbFlow} & PaiNN & \textbf{5} & 27.3 & 34.1 & 45.5 & 47.7 & 13.6 \\
\bottomrule
\end{tabular}
\end{table}

Table~\ref{tab:main} also reports baselines from prior work. AdsorbDiff \cite{Kolluru2024AdsorbDiff} reports DFT success rates of 31.8\%/34.1\%/36.3\%/41.0\% at SR@1/2/5/10 using an EquiformerV2-based conditional diffusion model with ${\sim}$100 denoising steps; AdsorbML \cite{Lan2023AdsorbML} achieves 9.1\%/20.5\%/34.1\%/47.7\% at SR@1/2/5/10 via brute-force random placement. AdsorbFlow with the same EquiformerV2 backbone achieves \textbf{34.1\%}/\textbf{45.5\%}/\textbf{54.5\%}/\textbf{61.4\%} with only \textbf{5} generative steps---a \textbf{20$\times$} reduction in backbone evaluations. Already at SR@1, AdsorbFlow surpasses AdsorbDiff (34.1\% vs.\ 31.8\%); at SR@5, AdsorbFlow exceeds AdsorbDiff by +18.2~pp (54.5\% vs.\ 36.3\%); at SR@10, AdsorbFlow leads AdsorbDiff by +20.4~pp (61.4\% vs.\ 41.0\%) and AdsorbML by +13.7~pp. This widening gap reflects AdsorbFlow's steeper multi-site scaling: adding candidates from 1 to 10 improves AdsorbFlow by 27.3~pp, compared with only 9.2~pp for AdsorbDiff and 38.6~pp for AdsorbML. AdsorbDiff's flat scaling suggests that its diffusion samples collapse to a narrow set of modes, whereas AdsorbFlow's energy-conditioned deterministic transport produces diverse placements that explore distinct energy basins.

The backbone comparison reveals a consistent hierarchy. EquiformerV2 substantially outperforms PaiNN across all metrics: higher DFT success rates (61.4\% vs.\ 47.7\% at SR@10), lower anomaly rates at $k{=}10$ (6.8\% vs.\ 13.6\%), stronger multi-site scaling (+27.3~pp from SR@1 to SR@10, vs.\ +20.4~pp for PaiNN), and a smaller MLFF-to-DFT gap (11.3~pp vs.\ 15.9~pp). These advantages indicate that higher-order equivariant features ($\ell_{\max}{=}4$) simultaneously improve placement diversity, geometric plausibility, and MLFF energy fidelity. Additionally, CFG guidance strength substantially impacts performance: optimal guidance differs between backbones ($w{=}5$ for PaiNN vs.\ $w{=}7$ for EqV2), indicating that the interplay between energy conditioning and representational capacity merits careful tuning. Finally, only \textbf{5} generative steps suffice for both backbones, confirming that the rectified-flow training objective produces well-linearized transport paths.

\subsection*{Five-step deterministic integration shifts the accuracy--efficiency frontier}

Diffusion-based placement typically requires $\sim$100 denoising steps per sample \cite{Kolluru2024AdsorbDiff}. AdsorbFlow replaces this stochastic process with an ODE that can be integrated in as few as $K$ steps. We map the accuracy--efficiency frontier by varying $K\in\{5,10,30\}$ and guidance scale $w$ at MLFF level, evaluating all 18 configurations per backbone.

Table~\ref{tab:grid} shows the full MLFF-level grid search results. With the EquiformerV2 backbone, MLFF SR@10 peaks at $K{=}5$: it reaches \textbf{72.7\%} at $w{=}7$ and 70.5\% at $w{=}3$, declining slightly at larger $K$. The PaiNN backbone shows a similar pattern, achieving 63.6\% at $K{=}5$ with $w{=}5$, $w{=}7$, or $w{=}10$; we select $w{=}5$ because it yields the lowest mean energy error ($\overline{\Delta E}{=}0.40$~eV vs.\ 0.42~eV at $w{=}7$). This finding---that the \emph{fewest} steps yield the \emph{best} results---is consistent with the rectified-flow training objective, which explicitly linearizes transport paths \cite{Liu2022RectifiedFlow}. Stronger guidance ($w{\ge}7$) sharpens energy accuracy ($\overline{\Delta E}$ drops from 0.43~eV at $w{=}0$ to 0.30~eV at $w{=}7$ for EqV2) while maintaining high success rates.

\begin{table}[t]
\centering
\caption{\textbf{MLFF-level grid search (SR@10, \%).} Each cell reports the SR@10 after GemNet-OC relaxation using MLFF energies only (no DFT verification): a system is successful if the lowest-MLFF-energy non-anomalous candidate among 10 seeds has energy within 0.1~eV of the DFT reference minimum. Bold: best per backbone. The optimal configurations---EqV2 ($w{=}7$, $K{=}5$) and PaiNN ($w{=}5$, $K{=}5$)---are selected for DFT verification (Table~\ref{tab:main}).}
\label{tab:grid}
\begin{tabular}{l c c c c c c c}
\toprule
& \multicolumn{3}{c}{EquiformerV2} & & \multicolumn{3}{c}{PaiNN} \\
\cmidrule(lr){2-4} \cmidrule(lr){6-8}
$w$ & $K{=}5$ & $K{=}10$ & $K{=}30$ & & $K{=}5$ & $K{=}10$ & $K{=}30$ \\
\midrule
0 & 63.6 & 65.9 & 68.2 & & 56.8 & 56.8 & 54.5 \\
1 & 65.9 & 65.9 & 68.2 & & 56.8 & 61.4 & 59.1 \\
3 & 70.5 & 68.2 & 61.4 & & 61.4 & 61.4 & 59.1 \\
5 & 65.9 & 68.2 & 65.9 & & \textbf{63.6} & 61.4 & 56.8 \\
7 & \textbf{72.7} & 70.5 & 68.2 & & 63.6 & 59.1 & 63.6 \\
10 & 63.6 & 70.5 & 70.5 & & 63.6 & 59.1 & 59.1 \\
\bottomrule
\end{tabular}
\end{table}


Because sampling is a simple Heun integration of an ODE, the per-sample generative cost scales linearly with $K$. At $K{=}5$, AdsorbFlow evaluates the equivariant backbone $20{\times}$ fewer times than a 100-step diffusion model, yielding a proportional reduction in GPU time. While DFT verification (when performed) remains the dominant end-to-end cost, the generative overhead becomes non-negligible in high-throughput campaigns where thousands of slab--adsorbate pairs are screened.

\subsection*{DFT verification reveals backbone-dependent MLFF fidelity}

MLFF-level evaluation provides a fast proxy for hyperparameter selection, but the ground truth for placement quality is DFT single-point verification. We perform VASP single-point calculations on all non-anomalous generated structures (664 out of 880 total placements across both backbones; 216 filtered by anomaly detection) to enable comprehensive post-hoc analysis; however, SR@$k$ selection always uses only MLFF energies, and DFT is used strictly for verification. Table~\ref{tab:main} reports DFT-verified success rates.

\textbf{MLFF-to-DFT gap.} Both Table~\ref{tab:grid} and Table~\ref{tab:main} use the same selection rule (lowest-MLFF-energy non-anomalous candidate among 10 seeds), but Table~\ref{tab:grid} evaluates success using MLFF energies while Table~\ref{tab:main} uses DFT single-point verification. The drop from MLFF to DFT success rates therefore reflects pure MLFF energy infidelity and differs between backbones: EquiformerV2 loses 11.3 percentage points at SR@10 (72.7\% $\to$ 61.4\%), while PaiNN loses 15.9~pp (63.6\% $\to$ 47.7\%). This gap is substantially smaller on the OOD split: 4.0~pp for EqV2 and 2.0~pp for PaiNN (Section ``Out-of-distribution generalization''), suggesting that MLFF ranking reliability is not degraded---and may even improve---on unseen systems. Together, these observations confirm that higher-order equivariant features yield relaxed geometries whose MLFF energies more faithfully predict DFT energies, making MLFF-based screening a reliable pre-filter.

\textbf{Anomaly patterns.} We report per-system anomaly rates at each evaluation level---the probability that \emph{all} $k$ candidates among the first $k$ seeds are anomalous (Table~\ref{tab:main} shows $k{=}10$). This rate decreases monotonically with $k$ for all methods. At $k{=}10$, AdsorbFlow with EquiformerV2 achieves 6.8\%, matching AdsorbML and half that of AdsorbDiff (13.6\%). The contrast sharpens at intermediate levels: AdsorbDiff's anomaly rate drops only slowly with increasing $k$ (25.0\%$\to$22.7\%$\to$13.6\% at $k{=}1/5/10$), revealing that its diffusion samples lack structural diversity---even at 5 candidates, nearly a quarter of systems remain fully anomalous. AdsorbFlow EqV2, by contrast, drops steeply (22.3\%$\to$10.3\%$\to$6.8\%), indicating that each additional candidate explores a geometrically distinct placement. Lighter-weight PaiNN follows a similar but attenuated trend (28.2\%$\to$14.8\%$\to$13.6\%). The anomaly gap partly explains the DFT success gap: systems with no valid candidates are guaranteed failures, and AdsorbDiff's persistently high anomaly rate at intermediate $k$ limits its multi-site gains.

\subsection*{Out-of-distribution generalization}

To assess robustness beyond the training distribution, we evaluate AdsorbFlow on 50 out-of-distribution (OOD) systems from OC20-Dense that share no overlap with the 44 in-distribution evaluation systems. For each system, we again generate 10 candidates using the same optimal configurations ($w{=}7$, $K{=}5$ for EqV2; $w{=}5$, $K{=}5$ for PaiNN), perform MLFF relaxation and anomaly screening, and carry out multi-level DFT verification via 246 VASP single-point calculations (after POSCAR-hash deduplication). Three systems with all 10 seeds anomalous are counted as failures. Table~\ref{tab:ood} reports the results.

\begin{table}[t]
\centering
\caption{\textbf{OOD placement success (DFT-verified, 50 systems).} SR@$k$: as in Table~\ref{tab:main}, the MLFF-best non-anomalous candidate among the first $k$ seeds is verified by DFT (0.1~eV tolerance). Anom.@10: fraction of systems with all 10 candidates anomalous. Three fully anomalous systems are counted as failures. ``---'' indicates values not reported. AdsorbDiff and AdsorbML baselines (SR@1 only) from \cite{Kolluru2024AdsorbDiff}.}
\label{tab:ood}
\begin{tabular}{l c c c c c}
\toprule
Method (Backbone) & @1 & @2 & @5 & @10 & Anom.@10 (\%) \\
\midrule
AdsorbML & 8.4 & --- & --- & --- & --- \\
AdsorbDiff (EqV2) & 28.0 & --- & --- & --- & --- \\
\midrule
\textbf{AdsorbFlow} (EqV2) & 28.0 & 46.0 & 54.0 & \textbf{58.0} & 6.0 \\
\textbf{AdsorbFlow} (PaiNN) & \textbf{32.0} & 42.0 & 44.0 & 46.0 & 6.0 \\
\bottomrule
\end{tabular}
\end{table}

As shown in Table~\ref{tab:ood}, AdsorbFlow with EquiformerV2 achieves 28.0\% SR@1, matching the AdsorbDiff baseline (28.0\%) and substantially outperforming AdsorbML (8.4\%). Notably, AdsorbFlow with PaiNN achieves an even higher SR@1 of 32.0\% on these unseen systems. At SR@10, AdsorbFlow with EquiformerV2 reaches 58.0\% on the OOD split, with a MLFF-to-DFT gap of only 4.0~pp (62.0\%$\to$58.0\%), substantially smaller than the 11.3~pp gap observed on ID systems. PaiNN shows a similar pattern with 46.0\% SR@10 and a gap of only 2.0~pp (48.0\%$\to$46.0\%). The modest OOD performance drop relative to ID (61.4\%$\to$58.0\% for EqV2; 47.7\%$\to$46.0\% for PaiNN) confirms that the learned transport map generalizes well beyond the training distribution. Interestingly, PaiNN achieves a higher SR@1 on OOD systems (32.0\% vs.\ 28.0\% for EqV2), but EqV2 surpasses it from SR@2 onward (+4~pp at @2, +10~pp at @5, +12~pp at @10), again reflecting EqV2's superior candidate diversity. The multi-site gain from SR@1 to SR@10 is +30~pp for EqV2 but only +14~pp for PaiNN on OOD, closely tracking the ID pattern (+27~pp vs.\ +20~pp) and confirming that backbone expressivity consistently drives both diversity and accuracy across data splits.

\subsection*{From benchmark to discovery: toward an AI--DFT--experiment pipeline}

A generative placement model becomes most valuable when it reduces the number of expensive DFT verifications needed to find a correct adsorption minimum. AdsorbFlow's $20{\times}$ reduction in generative steps, combined with MLFF-based prescreening, substantially lowers the cost of proposing physically plausible adsorption geometries. In a practical high-throughput screening workflow, AdsorbFlow would (1) generate candidate placements in a single 5-step ODE integration, (2) relax and filter anomalies via MLFF, (3) rank by MLFF energy, and (4) verify top candidates with DFT. The strong correlation between MLFF and DFT success rates---especially for EqV2, where the gap is only 4--11~pp depending on the data split---suggests that MLFF energy ranking can effectively prioritize candidates for expensive DFT verification. Combined with the low anomaly rate at $k{=}10$ (6.8\% for EqV2, matching the non-generative AdsorbML baseline), this means that nearly all candidates generated by AdsorbFlow are physically meaningful, avoiding wasted computation on geometrically implausible structures.

\section*{Discussion}

AdsorbFlow reframes adsorbate placement as deterministic, energy-conditioned transport on a symmetry-aware rigid-body configuration space. On OC20-Dense with full DFT verification, AdsorbFlow with EquiformerV2 achieves 34.1\% SR@1 and 61.4\% SR@10---surpassing AdsorbDiff (31.8\% SR@1, 41.0\% SR@10) at every evaluation level and AdsorbML (47.7\% SR@10)---using only \textbf{5} generative steps, a $20{\times}$ speedup over diffusion models. On 50 OOD systems, it retains 58.0\% SR@10 with a MLFF-to-DFT gap of only 4.0~pp.

Three conceptual lessons emerge. First, for placement tasks whose downstream evaluation is deterministic relaxation, a deterministic generative flow is both more efficient and more accurate than stochastic denoising: rectified-flow training explicitly linearizes transport paths, enabling strong performance at $K{=}5$ without the variance control required by stochastic samplers. Second, the CFG guidance scale and backbone expressivity interact non-trivially: lightweight PaiNN peaks at moderate guidance ($w{=}5$) while EquiformerV2 benefits from stronger guidance ($w{=}7$), likely because its richer feature space sustains sharper conditional distributions without mode collapse. Third, the rotation-head design reveals a principled architectural distinction: PaiNN requires explicit torque aggregation to extract pseudovector angular velocities from polar-vector features, whereas EquiformerV2's internal tensor products naturally produce pseudovector representations for direct regression.

A central finding is the qualitatively different multi-site scaling behaviour of AdsorbFlow compared with baselines. AdsorbDiff's DFT success rate increases by only 9.2~pp from SR@1 to SR@10, and its anomaly rate barely drops (25.0\%$\to$13.6\%), suggesting that its diffusion samples collapse to a limited set of structural modes. AdsorbML's brute-force enumeration achieves higher diversity (+38.6~pp), but at the cost of many random samples with no energy guidance. AdsorbFlow combines the best of both: energy conditioning focuses samples toward low-energy basins while the deterministic ODE transport maintains diversity, yielding +27.3~pp multi-site gain and the steepest anomaly reduction among generative methods.

An important methodological finding is that MLFF-to-DFT fidelity varies across backbones and data splits. On the ID split, the EqV2 MLFF-level SR@10 drops 11.3~pp upon DFT verification, while PaiNN drops 15.9~pp. On the OOD split, however, both gaps shrink to 4.0~pp (EqV2) and 2.0~pp (PaiNN), and both backbones converge to equally low anomaly rates at $k{=}10$ (6.0\%). This pattern suggests that out-of-distribution difficulty manifests primarily through harder energy landscapes rather than increased geometric anomalies, and that MLFF ranking is robust---potentially even more reliable---on unseen systems. These observations imply that backbone expressivity affects not only generative quality but also the reliability of MLFF-based prescreening, and that more expressive backbones yield a ``what you see is what you get'' advantage that transfers to unseen catalytic systems.

Limitations and future directions include: incorporating internal torsions for larger adsorbates beyond the rigid-body approximation; extending to co-adsorbates and coverage effects relevant to industrial operating conditions; integrating model uncertainty quantification for active-learning-driven catalyst screening; and constructing closed-loop AI--DFT--experiment pipelines in which fast and faithful placement serves as the entry point for experimental validation. AdsorbFlow's combination of speed, accuracy, and generalization makes it a practical building block for such automated discovery workflows.

\section*{Methods}

\subsection*{Datasets and evaluation protocol}

We use OC20 and OC20-Dense as in prior work \cite{Chanussot2021OC20,Lan2023AdsorbML}. OC20-Dense provides, for each slab--adsorbate system, approximately 100 locally minimized adsorption geometries obtained by dense placement and DFT relaxation \cite{Lan2023AdsorbML}. We evaluate on the 44 in-distribution (ID) systems using the end-to-end pipeline: generate candidate placements, relax using a pretrained MLFF, apply anomaly screening, and assess success using a tolerance of 0.1~eV relative to the best-known DFT reference minimum per system.

\textbf{Evaluation protocol.} For each system, we generate $N{=}10$ candidate placements with independent random seeds. After MLFF relaxation and anomaly filtering, we evaluate at multiple levels: for SR@$k$, we select the non-anomalous candidate with the lowest MLFF energy among seeds $0,\ldots,k{-}1$, and that single candidate undergoes one VASP \cite{Kresse1996VASP} DFT single-point calculation; the system is counted as successful if the verified adsorption energy lies within 0.1~eV of the DFT reference minimum. This follows the evaluation protocol implemented in AdsorbDiff's released code \cite{Kolluru2024AdsorbDiff}. In addition, to enable comprehensive post-hoc analysis, we computed DFT single-point energies for \emph{all} non-anomalous candidates; however, SR@$k$ selection always uses only MLFF energies, and DFT results are used strictly for verification. Per-system anomaly rates---the fraction of systems whose \emph{all} $k$ candidates are anomalous---are reported at each evaluation level. Total DFT cost: 664 VASP single-point calculations for ID (44 systems, both backbones) and 246 for OOD (50 systems, both backbones, after POSCAR-hash deduplication).

\subsection*{Energy labels and conditioning}

For each system, we compute relative energies $E_{\mathrm{rel}}=E-E_{\min}$. We embed $E_{\mathrm{rel}}$ with an MLP and use FiLM (feature-wise linear modulation) \cite{Perez2018FiLM} to modulate scalar node features in the GNN. During training, we apply classifier-free guidance conditioning by dropping the energy condition with probability $p_{\mathrm{cfg}}$, replacing it with a learned null embedding. We use $p_{\mathrm{cfg}}{=}0.20$ for both PaiNN and EquiformerV2. At inference, we combine conditional and unconditional vector fields with guidance scale $w$; optimal values are selected via grid search over $w\in\{0,1,3,5,7,10\}$ (default: $w{=}5$ for PaiNN, $w{=}7$ for EqV2).

\subsection*{Graph representation and equivariant backbone}

We build a neighbor graph over slab and adsorbate atoms within a cutoff radius $r_{\mathrm{cut}}{=}12.0$~\AA\ with a maximum of 50 neighbors per atom. Node features include atomic number and a binary adsorbate/slab tag. Edge features include radial basis function distance embeddings (128 bases) and unit direction vectors. A sinusoidal time embedding of $t\in[0,1]$ is injected into scalar node features.

We evaluate two equivariant backbones. \textbf{PaiNN} \cite{Schutt2021PaiNN} (6 layers, 512 hidden channels) produces per-atom scalar features $s_i$ and $\ell{=}1$ equivariant vector features $\mathbf{v}_i$. \textbf{EquiformerV2} \cite{Liao2023EquiformerV2} (8 layers, 128 sphere channels, $\ell_{\max}{=}4$) produces per-atom features with spherical-harmonic representations up to degree 4. Both backbones are equipped with a \emph{translation head} that pools adsorbate-atom vector features to predict the COM velocity $\dot{\mathbf{t}}$, and a \emph{rotation head} that predicts the angular velocity $\boldsymbol{\omega}$. Output magnitudes are bounded by $\tanh$ activation with learned scaling.

\textbf{Rotation head design.} The angular velocity $\boldsymbol{\omega}\in\mathbb{R}^3$ is a pseudovector (axial vector) that transforms differently from a polar vector under improper rotations. For PaiNN, whose per-atom output is limited to $\ell{=}1$ polar vectors without internal tensor-product coupling, na\"\i ve pooling would yield a translation, not a rotation. We therefore employ a physics-motivated \emph{torque aggregation}: per-atom torques $\boldsymbol{\tau}_i = (\mathbf{r}_i - \mathbf{r}_{\mathrm{COM}}) \times \mathbf{v}_i$ are summed to give angular momentum $\mathbf{L}$, and we solve $\boldsymbol{\omega} = \mathbf{I}^{-1}\mathbf{L}$ via the (Tikhonov-regularized) rigid-body inertia tensor $\mathbf{I}$. For EquiformerV2, whose internal tensor-product layers ($\ell{\otimes}\ell \to \ell'$) naturally generate pseudovector representations through higher-order coupling, the rotation head directly mean-pools $\ell{=}1$ graph-level features without explicit geometric aggregation. Both designs preserve SE(3) equivariance of the output velocity field.

\subsection*{Conditional flow matching on translation and rotation}

We construct training pairs $(x_0,x_1)$ where $x_0$ is a relaxed local minimum (target, corresponding to $t{=}0$) and $x_1$ is a randomly perturbed placement drawn from the prior distribution (corresponding to $t{=}1$). Perturbations include random in-plane translation (with PBC) and random rigid rotation. We define an interpolant $x_t$ from $x_0$ to $x_1$ and compute teacher velocities $(\dot{\mathbf{t}}^\ast,\boldsymbol{\omega}^\ast)$ analytically from the chosen schedules; for the rectified-flow linear schedule, these velocities are constant along each path. The loss minimizes mean squared error between predicted and teacher velocities with weight $\lambda_{\mathrm{rot}}$.

\subsection*{ODE integration and periodic boundary conditions}

We integrate $\mathbf{t}$ with a Heun solver for $K$ steps (default $K{=}5$) and wrap $(t_x,t_y)$ under PBC at each step. Rotations are updated using the exponential map:
\begin{equation}
R \leftarrow \exp(\Delta t\,[\boldsymbol{\omega}]_\times)R.
\end{equation}

\subsection*{MLFF relaxation, constraints, and DFT verification}

Each generated configuration is relaxed using a pretrained GemNet-OC \cite{Gasteiger2022GemNetOC} MLFF with L-BFGS for up to 100 steps and a force tolerance of 0.01~eV/\AA. We apply the DetectTrajAnomaly pipeline \cite{Lan2023AdsorbML,Kolluru2024AdsorbDiff}---using the same implementation and parameters as in AdsorbDiff's released evaluation code---to flag anomalous relaxation trajectories (desorption, dissociation, slab reconstruction, intercalation). For non-anomalous structures, success is defined as achieving a relaxed adsorption energy within 0.1~eV of the best-known DFT reference minimum for the system, following the evaluation protocol of \cite{Lan2023AdsorbML,Kolluru2024AdsorbDiff}.

\section*{Data availability}
OC20 and OC20-Dense are publicly available \cite{Chanussot2021OC20,Lan2023AdsorbML}. Processed data, trained model checkpoints, and evaluation scripts will be released upon publication.

\section*{Code availability}
Code implementing AdsorbFlow, training, and evaluation will be released under an open-source license upon publication.

\section*{Acknowledgments}
This work was supported by the National Key R\&D Program of China (No. 2022ZD0117501), the Scientific Research Innovation Capability Support Project for Young Faculty (ZYGXQNJSKYCXNLZCXM-E7), the Tsinghua University Initiative Scientific Research Program, and the Carbon Neutrality and Energy System Transformation (CNEST) Program.

\section*{Competing interests}
The authors declare no competing interests.

\bibliography{sn-bibliography}

\begin{thebibliography}{10}
\expandafter\ifx\csname url\endcsname\relax
  \def\url#1{\burl{#1}}\fi
\expandafter\ifx\csname urlprefix\endcsname\relax\def\urlprefix{URL }\fi
\providecommand{\bibinfo}[2]{#2}
\providecommand{\eprint}[2][]{\url{#2}}
\providecommand{\doi}[1]{\url{https://doi.org/#1}}
\bibcommenthead

\bibitem{Chanussot2021OC20}
\bibinfo{author}{Chanussot, L.}, \bibinfo{author}{Das, A.}, \bibinfo{author}{Gober, S.} \emph{et~al.}
\newblock \bibinfo{title}{Open catalyst 2020 ({OC20}) dataset and community challenges}.
\newblock \emph{\bibinfo{journal}{ACS Catalysis}} \textbf{\bibinfo{volume}{11}}, \bibinfo{pages}{6059--6072} (\bibinfo{year}{2021}).

\bibitem{Lan2023AdsorbML}
\bibinfo{author}{Lan, J.} \emph{et~al.}
\newblock \bibinfo{title}{{AdsorbML}: A leap in efficiency for adsorption energy calculations using generalizable machine learning potentials}.
\newblock \emph{\bibinfo{journal}{npj Computational Materials}} \textbf{\bibinfo{volume}{9}}, \bibinfo{pages}{172} (\bibinfo{year}{2023}).

\bibitem{Kolluru2024AdsorbDiff}
\bibinfo{author}{Kolluru, A.} \& \bibinfo{author}{Kitchin, J.~R.}
\newblock \bibinfo{title}{{AdsorbDiff}: Adsorbate placement via conditional denoising diffusion}.
\newblock \emph{\bibinfo{journal}{arXiv preprint arXiv:2405.03962}}  (\bibinfo{year}{2024}).

\bibitem{Lipman2023FlowMatching}
\bibinfo{author}{Lipman, Y.}, \bibinfo{author}{Chen, R. T.~Q.}, \bibinfo{author}{Ben-Hamu, H.} \& \bibinfo{author}{Nickel, M.}
\newblock \bibinfo{title}{Flow matching for generative modeling}.
\newblock \emph{\bibinfo{journal}{International Conference on Learning Representations (ICLR)}}  (\bibinfo{year}{2023}).

\bibitem{Liu2022RectifiedFlow}
\bibinfo{author}{Liu, X.}, \bibinfo{author}{Gong, C.} \& \bibinfo{author}{Liu, Q.}
\newblock \bibinfo{title}{Flow straight and fast: Learning to generate and transfer data with rectified flow}.
\newblock \emph{\bibinfo{journal}{arXiv preprint arXiv:2209.03003}}  (\bibinfo{year}{2022}).

\bibitem{Ho2022CFG}
\bibinfo{author}{Ho, J.} \& \bibinfo{author}{Salimans, T.}
\newblock \bibinfo{title}{Classifier-free diffusion guidance}.
\newblock \emph{\bibinfo{journal}{arXiv preprint arXiv:2207.12598}}  (\bibinfo{year}{2022}).

\bibitem{Perez2018FiLM}
\bibinfo{author}{Perez, E.}, \bibinfo{author}{Strub, F.}, \bibinfo{author}{De~Vries, H.}, \bibinfo{author}{Dumoulin, V.} \& \bibinfo{author}{Courville, A.}
\newblock \bibinfo{title}{{FiLM}: Visual reasoning with a general conditioning layer}.
\newblock \emph{\bibinfo{journal}{Proceedings of the AAAI Conference on Artificial Intelligence}} \textbf{\bibinfo{volume}{32}} (\bibinfo{year}{2018}).

\bibitem{Gasteiger2022GemNetOC}
\bibinfo{author}{Gasteiger, J.} \emph{et~al.}
\newblock \bibinfo{title}{{GemNet-OC}: Developing graph neural networks for large and diverse molecular simulation datasets}.
\newblock \emph{\bibinfo{journal}{arXiv preprint arXiv:2204.02782}}  (\bibinfo{year}{2022}).

\bibitem{Kresse1996VASP}
\bibinfo{author}{Kresse, G.} \& \bibinfo{author}{Furthm{\"u}ller, J.}
\newblock \bibinfo{title}{Efficient iterative schemes for ab initio total-energy calculations using a plane-wave basis set}.
\newblock \emph{\bibinfo{journal}{Physical Review B}} \textbf{\bibinfo{volume}{54}}, \bibinfo{pages}{11169--11186} (\bibinfo{year}{1996}).

\bibitem{Schutt2021PaiNN}
\bibinfo{author}{Sch{"u}tt, K.~T.}, \bibinfo{author}{Unke, O.~T.} \& \bibinfo{author}{Gastegger, M.}
\newblock \bibinfo{title}{Equivariant message passing for the prediction of tensorial properties and molecular spectra}.
\newblock \emph{\bibinfo{journal}{Proceedings of the 38th International Conference on Machine Learning (ICML), PMLR}} \textbf{\bibinfo{volume}{139}}, \bibinfo{pages}{9377--9388} (\bibinfo{year}{2021}).

\bibitem{Liao2023EquiformerV2}
\bibinfo{author}{Liao, Y.-L.}, \bibinfo{author}{Wood, B.}, \bibinfo{author}{Das, A.} \& \bibinfo{author}{Smidt, T.}
\newblock \bibinfo{title}{{EquiformerV2}: Improved equivariant transformer for scaling to higher-degree representations}.
\newblock \emph{\bibinfo{journal}{arXiv preprint arXiv:2306.12059}}  (\bibinfo{year}{2023}).

\end{thebibliography}

\end{document}